\documentclass{egpubl}
\usepackage{vcbm2024s}

 \WsShortPaper      

\usepackage[T1]{fontenc}
\usepackage{dfadobe}  

\usepackage{cite} 
\BibtexOrBiblatex
\electronicVersion
\PrintedOrElectronic
\ifpdf \usepackage[pdftex]{graphicx} \pdfcompresslevel=9
\else \usepackage[dvips]{graphicx} \fi

\usepackage{egweblnk} 
\usepackage{subcaption}
\usepackage{amsmath}
\usepackage{siunitx}
\usepackage{float}
\DeclareSIUnit{\molar}{M}
\DeclareSIUnit{\literklein}{l}

\title[Quantification of \textit{in vitro} Wound Healing Scratch Assays]%
      {Virtually Objective Quantification of \textit{in vitro} Wound Healing Scratch Assays with the Segment Anything Model}

\author[K. Löwenstein, J. Rehrl, A. Schuster \& M. Gadermayr]
{\parbox{\textwidth}{\centering Katja Löwenstein$^{1}$,
         Johanna Rehrl$^{2,3}$,
         Anja Schuster$^{2}$ and
         Michael Gadermayr$^{1}$
        }
        \\
{\parbox{\textwidth}{\centering $^1$Department of Information Technology and Digitalisation, Salzburg University of Applied Sciences, Salzburg, Austria\\
         $^2$Department of Health Sciences, Salzburg University of Applied Sciences, Salzburg, Austria\\
         $^3$Department of Biosciences and Medical Biology, Paris-Lodron University of Salzburg, Hellbrunner Straße 34, 5020 Salzburg, Austria
       }
}
}
\begin{document}

\maketitle
\begin{abstract}
   The \textit{in vitro} scratch assay is a widely used assay in cell biology to assess the rate of wound closure related to a variety of therapeutic interventions. While manual measurement is subjective and vulnerable to intra- and interobserver variability, computer-based tools are theoretically objective, but in practice often contain parameters which are manually adjusted (individually per image or data set) and thereby provide a source for subjectivity. Modern deep learning approaches typically require large annotated training data which complicates instant applicability. In this paper, we make use of the segment anything model, a deep foundation model based on interactive point-prompts, which enables class-agnostic segmentation without tuning the network's parameters based on domain specific training data.
   The proposed method clearly outperformed a semi-objective baseline method that required manual inspection and, if necessary, adjustment of parameters per image. 
   Even though the point prompts of the proposed approach are theoretically also a source for subjectivity, results attested very low intra- and interobserver variability, even compared to manual segmentation of domain experts.\\

\end{abstract}  
 
\section{Motivation}
Quantitative assessment of the rate of wound closure using \textit{in vitro} scratch assay is essential since it enables precise, consistent and reproducible measurement of cell migration and proliferation rates, facilitating the evaluation of potential therapeutic interventions or drug effects on wound closure~\cite{Stamm2016}. Even though the \textit{in vitro} scratch assay has a number of limitations in comparison to 3D cell culture  or \textit{in vivo} models, it is still the most frequently used technique due to its benefits of being quick, reliable, and affordable \cite{Sarian2023}. By quantifying parameters such as wound area closure and cell migration distance, researchers can objectively analyse the efficacy of treatments and gain valuable insights into the underlying mechanisms of the wound healing process, such as  proliferation, migration, protein synthesis, cell-cell interaction, cell-matrix interaction, wound contraction, epithelialisation, tensile strength, and morphology.
Objective analysis of such parameters requires time-lapse microscopy in combination with the objective analysis of the digital image data.
Image analysis in this setting typically refers to capturing the wound boarders based on image segmentation.
Manual assessment (i.e. manually creating polygons) of such image data is subjective, error-prone and time-consuming and prevents objective large scale studies~\cite{Joskowicz2018}.
Perfectly objective analysis requires computer-based methods without any parameters to adjust. Depending on the biological and the imaging setting, however, images show high variability with the potential of affecting computer-based image analysis methods. The solution to circumvent degraded accuracy of image analysis is to provide adjustment parameters (which can be used to manually tweak the output). Particularly modern deep learning based methods combined with domain specific data augmentation are theoretically able to cope with large variability in the data (even without any parameters to adjust). Such methods, however, typically require large amounts of training data that cover variability. Data augmentation is a method to reduce the need for large data, nevertheless, domain knowledge is needed to select the ideal augmentation methods.
Recently, there is strong focus on foundation models which enable data-efficient adaptation to novel tasks.
\begin{figure*}[t]
    \centering
    \includegraphics[width=0.99\linewidth]{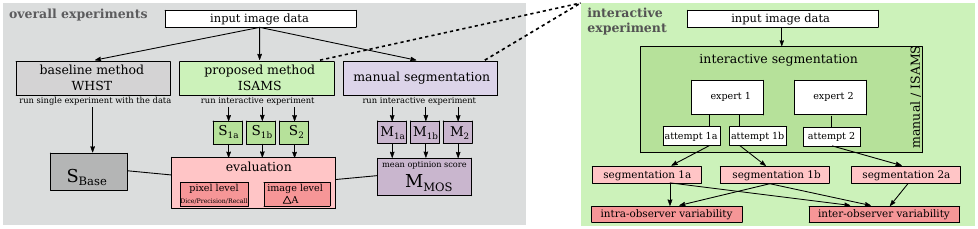}
    \caption{Overview of the experiments performed in this study. While the right box shows all sub-experiments performed for both interactive segmentations, manual and ISAMS, the left box shows the overall big picture including evaluation.}
    \label{fig:graphicalabstract}
\end{figure*}

In this work, we investigate an application agnostic approach, i.e. the method is neither (fine)tuned nor adjusted in any sense with respect to the application scenario or the underlying data. All domain specific input, specifically interactive prompts based on mouse clicks, is obtained from a user during interactive segmentation.
There already exist plenty of models for interactive prompt-based segmentation. The following three methods all have in common that they encode and process the additional user input and use existing semantic segmentation architectures as backbone.
RITM by Sofiiuk et al.~\cite{ritm} uses DeepLabV3+ and HRNet+OCR, Focalclick by Chen et al.~\cite{focalclick} also employ HRNet+OCR and SegFormer, and SimpleClick by Liu et al.~\cite{simpleclick} uses vision transformers.
A rather new method is the Segment Anything Model (SAM)~\cite{kirillov2023segment} which is based on a state-of-the-art deep foundation model in combination with different types of prompt encoding. Compared to the previous approaches, inspired by language models, SAM performs prompt encoding based on tokens.
Mazurowski et al. ~\cite{Mazurowski2023} conducted a study on 19 datasets to investigate the performance of SAM applied to medical images, compared to the three previously mentioned methods. Independently of the data set, SAM outperformed all other methods.
To optimize the method for processing medical imaging, derivatives trained on domain specific data were proposed~\cite{medsam,Archit2023}.

\noindent \textbf{Contribution:} In this work we investigated an interactive deep foundation model, specifically the so-called segment anything model (SAM), trained on large, general purpose training data. This model was directly applied to the generated microscopic wound healing image data in combination with point prompts obtained by domain experts. We analysed the performance compared to a baseline method and the effect of varying input prompts obtained by different users on the final segmentation performance. We particularly investigated the effect of intra- and interobserver variability on the overall performance. For that purpose, we acquired and analysed a ground truth based on 3 manual observations.
All data as well as the source code will be publicly provided upon acceptance.


\section{Materials and Methods}

In a nutshell, we propose a workflow and evaluate the performance utilizing a prompt-based interactive segmentation model (section~\ref{subsec:interactive_segmentation}) for segmenting wound healing scratch assays. On top of a conventional comparison with a manual ground truth and a baseline method from literature we performed repeated runs to assess intra- and interobserver variability (section\mbox{s~\ref{subsec:groundtruth} - \ref{subsec:baseline_method}}). An overview of the experiments is provided in Figure~\ref{fig:graphicalabstract}.

\subsection{Interactive Prompt-based Segmentation}
\label{subsec:interactive_segmentation}
We propose the Interactive SAM-based Segmentation (ISAMS) framework based on the Segment Anything Model (SAM)~\cite{kirillov2023segment} for interactively segmenting wound healing images. SAM is a promptable segmentation method trained on a large, general-purpose dataset and comprises an image encoder, a prompt encoder, and a mask decoder. The method efficiently processes input images and prompts to generate binary segmentation masks in real time. Prompts can be made by positive and negative mouse clicks, or based on bounding boxes. Even though the models can be optimized for certain domains (e.g. for medical image data~\cite{medsam} or microscopy data~\cite{Archit2023}), we made use of the generic model trained on general purpose image data, since we aim to assess the direct (training-free) and generic applicability of the approach.
In the proposed setting, first a user clicks on a point inside the wound area and immediately obtained feedback in the form of a segmentation overlay. Depending if the segmentation is already sufficiently accurate, the user has the option for performing arbitrary additional positive clicks inside the wound and/or negative clicks outside the wound.
We utilized the user interface implementation of SAM-webui (\url{https://github.com/derekray311511/SAM-webui}) for segmentation and mask generation. 

\subsection{Ground Truth Annotation \& Mean Opinion Score}
\label{subsec:groundtruth}
For performing manual annotations, we employed Labelme~\cite{wada2018labelme} to acquire a ground truth based on polygonal annotations. To reduce the subjective component, we collected three manual segmentation sets obtained from two different domain experts. Thereby, three annotations were generated for each image in the data set. The final reference segmentation mask (for evaluating the automated appraoches) was created based on the mean opinion score. For that purpose, we performed a majority voting on pixel level, i.e. each pixel in each segmented image was categorized as either "wound" or "background" i.e. cell covered area. In the resulting mean image, the majority opinion for each pixel was computed.

\begin{figure*}[h!]
    \centering
    \includegraphics[width=.99\textwidth]{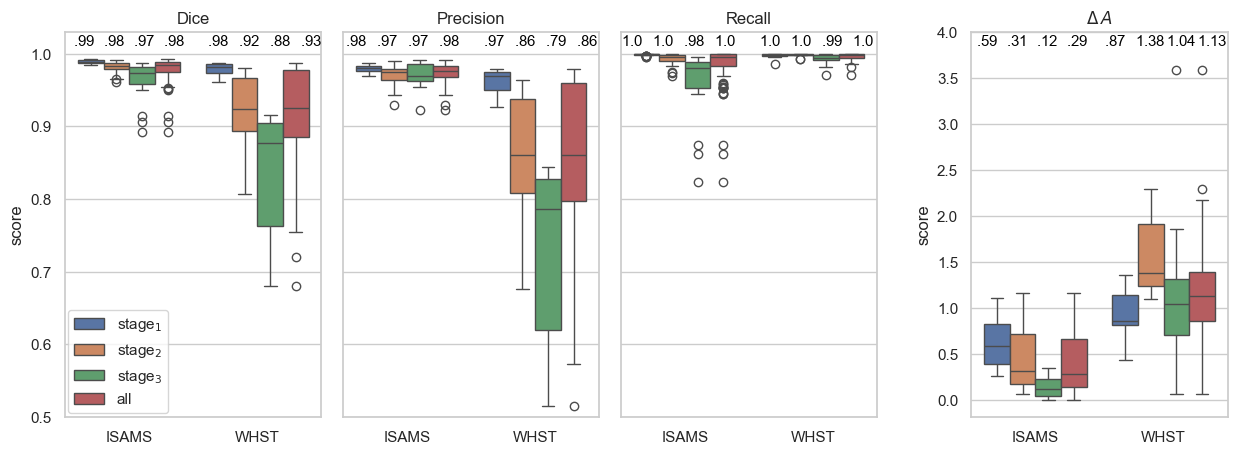}
    \caption{\label{fig:subplot-4metrics}
           Results of the methods ISAMS and WHST based on the metrics Dice, precision, recall and $\Delta A$.}
\end{figure*}

\begin{figure*}[h!]
    \centering
    \includegraphics[width=.99\textwidth]{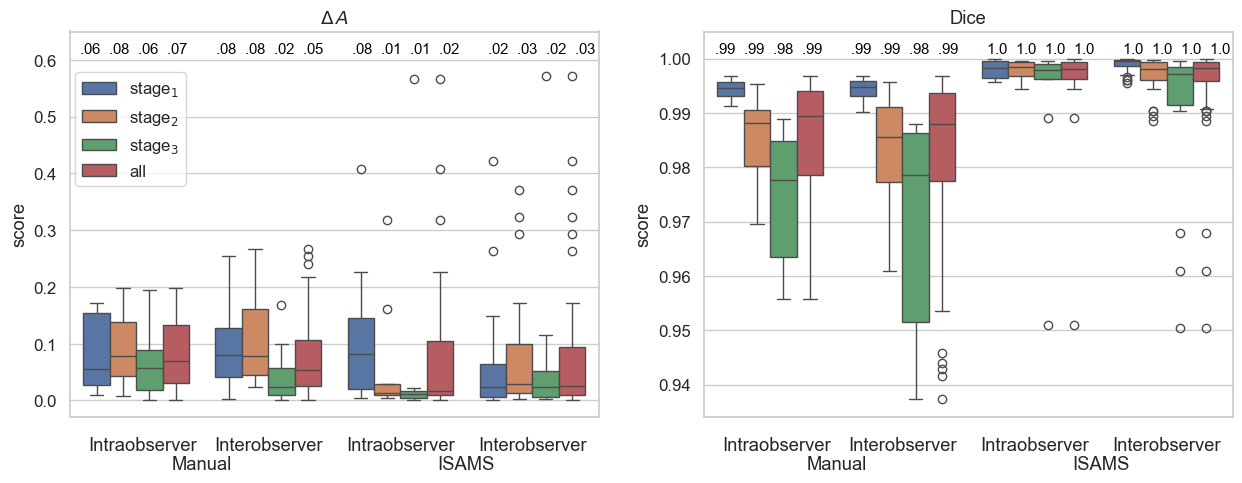}
    \caption{\label{fig:subplot-2metrics}
           Comparison of intra- and interobserver variability based on the measures $\Delta A$ and Dice for manual segmentation and ISAMS.}
\end{figure*}

\subsection{Evaluation Metrics}
For evaluation, we used Dice similarity coefficient, precision, recall and the absolute area difference.
The latter one is represented by the difference between the areas of two segmentations ($A_1$, $A_2$, one of them is typically a ground truth) in relation to the total area of the image in pixels $A_{total}$, which is calculated as $\Delta A = \frac{\left| A_1 - A_2 \right|}{A_{total}}$.

Dice and $\Delta$A were also utilized to assess intra- and interobserver variability. 
In this setting an arbitrary number of segmentation maps per image and per observer is available and not (as typically the case) a single pair consisting of a ground truth and a test sample (see Figure~\ref{fig:graphicalabstract}). 
To facilitate the use of these two measures to quantify intra- and interobserver variability, we sampled all subsets of size 2 within individual observers (intraobserver variability) and all subsets of size 2 between observers (interobserver variability).
Finally the value of all paired comparisons were aggregated.

\subsection{Baseline Method}
\label{subsec:baseline_method}
The Wound Healing Scoring Tool (WHST) proposed by Suarez-Arnedo et al.~\cite{SuarezArnedo2020} exhibits a well-studied method dedicated to the specific application scenario. We made use of the reference ImageJ implementation of WHST. It was designed for quantifying wound and cell covered areas in wound healing assays based on classical segmentation algorithms focusing on neighboring pixel intensity variance to distinguish between cell monolayer and open wound areas. The algorithm enhances image contrast, applies a variance filter and performs morphological reconstruction to refine the segmentation. The method offers customizable parameters for variance filter radius, binarization threshold and contrast enhancement. For our purpose, the parameter selection was determined by trial and error and set to a threshold value of 100 and a percentage of saturated pixels to 0.4. The variance window radius was set to 10 and reduced if necessary depending on the microscope settings used for image acquisition.

\begin{figure*}[h!]
    \centering
    \includegraphics[width=.99\textwidth]{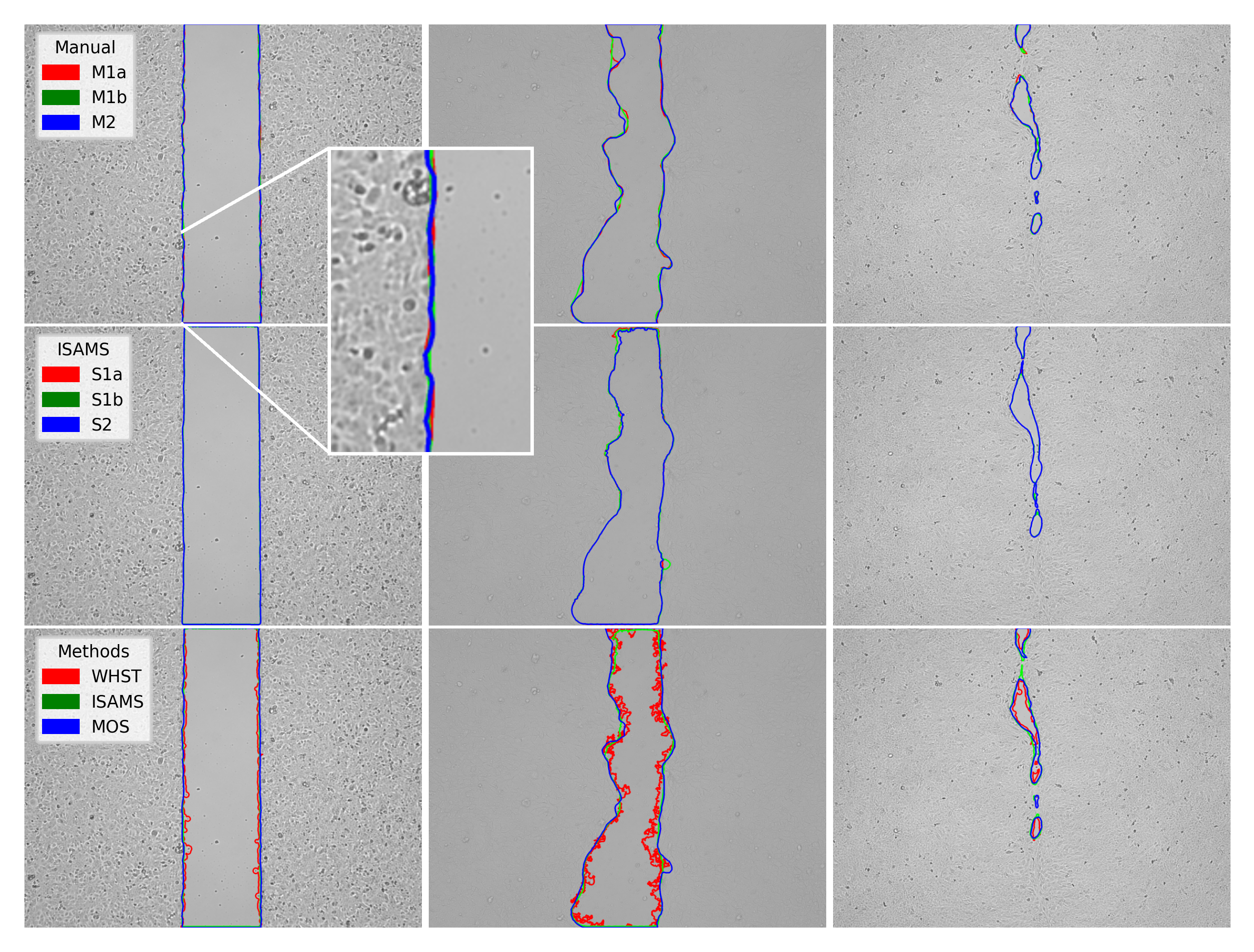}
    \caption{\label{fig:mosaik}
           Differences in wound segmentation between experts during manual segmentation (first row), during interactive segmentation using ISAMS (second row) and between the different methods (third row) at three stages (see different columns)}
\end{figure*}

\subsection{Data Acquisition} 

HaCaT cells, a spontaneously immortalized human keratinocyte cell line, were cultured in Dulbecco’s Modified Eagle Medium (DMEM) high glucose (\SI{4.5}{\gram\per\literklein}) (Sigma-Aldrich, St. Louis, USA) supplemented with \SI{10}{\percent} fetal bovine serum (Capricorn, Ebsdorfergrund, Germany), \SI{1}{\percent} L-glutamine (\SI{200}{\milli\molar}) (Capricorn, Ebsdorfergrund, Germany), \SI{1}{\percent} penicillin-streptomycin (\SI[group-digits=true]{10000}{} units penicillin and \SI{10}{\milli\gram} streptomycin/ml) (Sigma-Aldrich, St. Louis, USA). They were maintained at \SI{37}{\celsius} and \SI{5}{\percent} CO$_{2}$ in a humid atmosphere. 
For the wound healing assay, HaCaT cells were seeded into two well silicone inserts with a \SI{500}{\micro\meter} cell free gap (ibidi, Gräfelfing, Germany) on 12-well cell culture plates (Greiner, Kremsmünster, Austria)  at a density of \SI{7.5e5}{cells\per\milli\literklein} (\SI{70}{\micro\literklein}/well). After 24 h of incubation, the inserts were removed and the confluent cell layers were washed twice with \SI{1}{\milli\literklein} sterile phosphate-buffered saline (PBS) (Gibco, Life Technologies, Carlsbad, USA). Subsequently, \SI{1}{\milli\literklein} of DMEM and \SI{1}{\milli\literklein} of tree bark extracts suspected to accelerate wound closure, or DMEM in the case of untreated cells, were added. The cell-free gaps were imaged at various time points after treatment using an Axiovert~5 inverted microscope (Zeiss, Oberkochen, Germany) and the ZEISS ZEN 3.9 software at 5x and 10x magnification.

The dataset utilized for the experiments consists of 30 images which were taken from different wound healing experiments. Those were manually divided into three levels of temporal stages (named stage$_1$, stage$_2$ and stage$_3$), each containing ten images.
By providing all data, including raw images, manual annotations and source code, we actively invite other researchers to contribute to built up on our research.

\section{Results}
Figure \ref{fig:subplot-4metrics} shows the results for the comparison of both methods ISAMS and WHST against the MOS obtained with manual segmentation. The numbers indicate the median value for each box. ISAMS shows clearly and consistently improved scores for the metrics $\Delta A$, Dice and precision, independently of the data set. For all metrics apart from $\Delta A$, particularly with WHST, the score decreases with later temporal stages. Recall is consistently high with both methods.

Figure \ref{fig:subplot-2metrics} exhibits quantitative results for intra- and interobserver variability for manual annotations and for segmentation with ISAMS, both individually for the three temporal groups and for the groups together. The plot on the left shows $\Delta A$, while the plot on the right displays the Dice score. The numbers indicate the median value for each box.
The variability of $\Delta A$ between different trials remains for both methods and most images minimal, under 0.5\%. The variability measured by Dice increases with the temporal stage of the image for manual annotation, while it does not exhibit significant changes for ISAMS.
While ISAMS consistently shows a significantly lower median variability when looking at Dice, the values are similar when looking at $\Delta A$.
Figure \ref{fig:mosaik} visualizes qualitative results for the overall experiments as well as the interactive experiments (Figure~\ref{fig:graphicalabstract}). The columns represent the three temporal stages, while the rows stand for the type of experiment.
We also analyzed the user interaction with ISAMS, more precisely we measured how long it takes the users to perform wound segmentation with ISAMS. The time measurement started with the click to load the next image until the click to save the mask. \SI{85}{\percent} percent of the images were processed in under 5 seconds, while \SI{97}{\percent} percent were analyzed within 20 seconds or less. The categories only had a minor impact on the analysis time.

\section{Discussion}
The presented study focused on the quantitative assessment of wound closure rates utilizing an in vitro scratch assay. The proposed interactive method ISAMS performed well overall, not only on average compared to a baseline method but also with respect to intra- and interobserver variability compared to manual segmentation performed by trained experts.
By comparing ISAMS with WHST, we observed consistent improvements, independent of the temporal stage and the evaluation measure. 
To obtain best insights, we investigated a pixel-level (Dice) and an aggregated domain specific measure ($\Delta A$) which is based on the wound area which is an important feature for biological analysis. $\Delta A$, however, only reflects the size of the area such that false positive and false negatives on pixel-level can finally cancel out, which is particularly the case in manual annotations (showing similar $\Delta A$ scores as ISAMS). On pixel level (Dice) these errors are clearly visible (indicated by clearly lower Dice scores in the case of WHST). An example can be seen in a zoom on the segmentation lines in Figure \ref{fig:mosaik}.
The analysis of user interaction with ISAMS indicated efficient segmentation performance, with most images analysed within seconds. The categories did not influence the segmentation time significantly.
All these findings underscore the potential of ISAMS as a reliable and efficient tool for accurate wound segmentation, offering objective and reproducible results while minimizing effort.
In this preliminary work, we performed experiments with a small set of manual observations and also a rather small data set. In future work, we plan to perform a larger study based on a large image data set and additional baseline methods. The publicly provided data is thereby intended to be enlarged as well. The comparison between different SAM-models, particularly those dedicated to medical imaging data such as MedSAM~\cite{medsam} or MicroSAM~\cite{Archit2023}, would also give valuable insights.
To conclude, this work showed that fine-tuning-free interactive segmentation based on the segment anything model has the potential to be used for efficient, accurate and virtually objective quantification of in-vitro wound healing studies and present an alternative to fully-automated segmentation models based on individually trained models.
The method showed high segmentation performance compared to a baseline method with minor manual interaction and effort. On top of that, we showed that intraobserver and interobserver variability can be clearly improved compared to the manual annotation even by trained and motivated experts.

\bibliographystyle{eg-alpha-doi} 
\bibliography{egbibsample}

\newcommand{\etalchar}[1]{$^{#1}$}
\begin{thebibliography}{\uppercase{SATFC{\etalchar{*}}20}}

\bibitem[ANK{\etalchar{*}}23]{Archit2023}
\textsc{Archit A., Nair S., Khalid N., Hilt P., Rajashekar V., Freitag M., Gupta S., Dengel A., Ahmed S., Pape C.}:
\newblock Segment anything for microscopy.
\newblock URL: \url{http://dx.doi.org/10.1101/2023.08.21.554208}, \href {https://doi.org/10.1101/2023.08.21.554208} {\path{doi:10.1101/2023.08.21.554208}}.

\bibitem[CZZ{\etalchar{*}}22]{focalclick}
\textsc{Chen X., Zhao Z., Zhang Y., Duan M., Qi D., Zhao H.}:
\newblock Focalclick: Towards practical interactive image segmentation.
\newblock In \emph{Proceedings of the IEEE/CVF Conference on Computer Vision and Pattern Recognition (CVPR)} (June 2022), pp.~1300--1309.

\bibitem[JCCS18]{Joskowicz2018}
\textsc{Joskowicz L., Cohen D., Caplan N., Sosna J.}:
\newblock Inter-observer variability of manual contour delineation of structures in ct.
\newblock \emph{European Radiology 29}, 3 (Sept. 2018), 1391–1399.
\newblock URL: \url{http://dx.doi.org/10.1007/s00330-018-5695-5}, \href {https://doi.org/10.1007/s00330-018-5695-5} {\path{doi:10.1007/s00330-018-5695-5}}.

\bibitem[KMR{\etalchar{*}}23]{kirillov2023segment}
\textsc{Kirillov A., Mintun E., Ravi N., Mao H., Rolland C., Gustafson L., Xiao T., Whitehead S., Berg A.~C., Lo W.-Y., Dollár P., Girshick R.}:
\newblock Segment anything, 2023.
\newblock \href {http://arxiv.org/abs/2304.02643} {\path{arXiv:2304.02643}}.

\bibitem[LXBN23]{simpleclick}
\textsc{Liu Q., Xu Z., Bertasius G., Niethammer M.}:
\newblock Simpleclick: Interactive image segmentation with simple vision transformers, 2023.
\newblock \href {http://arxiv.org/abs/2210.11006} {\path{arXiv:2210.11006}}.

\bibitem[MDG{\etalchar{*}}23]{Mazurowski2023}
\textsc{Mazurowski M.~A., Dong H., Gu H., Yang J., Konz N., Zhang Y.}:
\newblock Segment anything model for medical image analysis: An experimental study.
\newblock \emph{Medical Image Analysis 89} (Oct. 2023), 102918.
\newblock URL: \url{http://dx.doi.org/10.1016/j.media.2023.102918}, \href {https://doi.org/10.1016/j.media.2023.102918} {\path{doi:10.1016/j.media.2023.102918}}.

\bibitem[MHL{\etalchar{*}}24]{medsam}
\textsc{Ma J., He Y., Li F., Han L., You C., Wang B.}:
\newblock Segment anything in medical images.
\newblock \emph{Nature Communications 15}, 1 (Jan. 2024).
\newblock URL: \url{http://dx.doi.org/10.1038/s41467-024-44824-z}, \href {https://doi.org/10.1038/s41467-024-44824-z} {\path{doi:10.1038/s41467-024-44824-z}}.

\bibitem[SATFC{\etalchar{*}}20]{SuarezArnedo2020}
\textsc{Suarez-Arnedo A., Torres~Figueroa F., Clavijo C., Arbeláez P., Cruz J.~C., Muñoz-Camargo C.}:
\newblock An image j plugin for the high throughput image analysis of in vitro scratch wound healing assays.
\newblock \emph{PLOS ONE 15}, 7 (July 2020), e0232565.
\newblock URL: \url{http://dx.doi.org/10.1371/journal.pone.0232565}, \href {https://doi.org/10.1371/journal.pone.0232565} {\path{doi:10.1371/journal.pone.0232565}}.

\bibitem[SPK22]{ritm}
\textsc{Sofiiuk K., Petrov I.~A., Konushin A.}:
\newblock Reviving iterative training with mask guidance for interactive segmentation.
\newblock In \emph{2022 IEEE International Conference on Image Processing (ICIP)} (2022), pp.~3141--3145.
\newblock \href {https://doi.org/10.1109/ICIP46576.2022.9897365} {\path{doi:10.1109/ICIP46576.2022.9897365}}.

\bibitem[SRS{\etalchar{*}}16]{Stamm2016}
\textsc{Stamm A., Reimers K., Strauß S., Vogt P., Scheper T., Pepelanova I.}:
\newblock In vitro wound healing assays – state of the art.
\newblock \emph{BioNanoMaterials 17}, 1–2 (Apr. 2016), 79–87.
\newblock URL: \url{http://dx.doi.org/10.1515/bnm-2016-0002}, \href {https://doi.org/10.1515/bnm-2016-0002} {\path{doi:10.1515/bnm-2016-0002}}.

\bibitem[SZZ{\etalchar{*}}23]{Sarian2023}
\textsc{SARIAN M.~N., ZULKEFLI N., ZAIN S., MANIAM S., FAKURAZI S.}:
\newblock A review with updated perspectives on in vitro and in vivo wound healing models.
\newblock \emph{Turkish Journal of Biology 47}, 4 (Aug. 2023), 236–246.
\newblock URL: \url{http://dx.doi.org/10.55730/1300-0152.2659}, \href {https://doi.org/10.55730/1300-0152.2659} {\path{doi:10.55730/1300-0152.2659}}.

\bibitem[Wad18]{wada2018labelme}
\textsc{Wada K.}:
\newblock labelme: Image polygonal annotation with python.
\newblock \url{https://github.com/wkentaro/labelme}, 2018.

\end{thebibliography}

\end{document}